\documentclass{amia}
\usepackage{graphicx}
\usepackage[labelfont=bf]{caption}
\usepackage[superscript,nomove]{cite}
\usepackage[dvipsnames]{xcolor}
\usepackage[sort&compress,comma,numbers,super]{natbib}
\usepackage{multirow}
\usepackage{url}
\usepackage{soul}
\usepackage{hyperref}
\usepackage{listings}
\usepackage{amsmath}
\usepackage{needspace}

\usepackage{booktabs}
\usepackage{multirow}
\usepackage{makecell}
\usepackage{array}
\usepackage{pifont}
\usepackage{bm}
\usepackage{relsize}
\usepackage{amsmath}
\usepackage{subcaption}
\usepackage{sidecap}

\newcommand\EatDot[1]{}

\definecolor{myblue}{HTML}{4E79A7}
\definecolor{mygreen}{HTML}{59A14F}
\definecolor{myred}{HTML}{E15759}

\newcommand{\tranx}{\mbox{\textsc{Tranx}}}
\newcommand{\cf}{\mbox{\textsc{Coarse2Fine}}}
\newcommand{\transformer}{\mbox{\textsc{Transformer}}}
\newcommand{\lexicon}{\mbox{\textsc{Lexicon-based}}}
\newcommand{\cmark}{\ding{51}}
\newcommand{\xmark}{\ding{55}}
\newcommand{\fhirdata}{\mbox{\textsc{FHIR\textsubscript{data}}}}
\newcommand{\icudata}{\mbox{\textsc{ICU\textsubscript{data}}}}

\begin{document}

\title{Toward a Neural Semantic Parsing System for EHR Question Answering}

\author{Sarvesh Soni, MS, Kirk Roberts, PhD}

\institutes{
    School of Biomedical Informatics \\
    The University of Texas Health Science Center at Houston \\
    Houston TX, USA \\
}

\maketitle

\noindent{\bf Abstract}

\textit{Clinical semantic parsing (SP) is an important step toward identifying the exact information need (as a machine-understandable logical form) from a natural language query aimed at retrieving information from electronic health records (EHRs).
Current approaches to clinical SP are largely based on traditional machine learning and require hand-building a lexicon.
The recent advancements in neural SP show a promise for building a robust and flexible semantic parser without much human effort.
Thus, in this paper, we aim to systematically assess the performance of two such neural SP models for EHR question answering (QA).
We found that the performance of these advanced neural models on two clinical SP datasets is promising given their ease of application and generalizability.
Our error analysis surfaces the common types of errors made by these models and has the potential to inform future research into improving the performance of neural SP models for EHR QA.
}

\section{Introduction}
\label{introduction}
\vspace{-5pt}

Navigating the information present in electronic health records (EHRs) is difficult due to various usability issues associated with these systems \cite{roman2017NavigationElectronicHealth}.
Question answering (QA) systems help in this regard by providing a way to express an information need in the form of natural language
(instead of through a long series of clicks).
One of the approaches to interpret and tackle the information need expressed by a natural language query is known as semantic parsing (SP) where the inherently ambiguous input query is mapped to an unambiguous logical form (LF) that is understandable by machines \cite{kamath2019SurveySemanticParsing}.
Most current approaches to clinical SP (where the queries are focused toward EHRs) have been centered around rule-based techniques and traditional machine learning (ML) \cite{patrick2012OntologyClinicalQuestions, roberts2017SemanticParsingMethod, soni2019UsingFHIRConstruct}.
Recently, many neural SP techniques have emerged that hold the promise to achieve comparable levels of performance as that of the traditional techniques while overcoming some of the difficulties involved in building the traditional systems.
However, it is unknown how these neural systems will perform in the clinical domain, given the domain-specific intricacies and difficulties such as the lack of large high-quality datasets.

Many traditional semantic parsers make use of a lexicon to map the phrases present in natural language (such as \textit{healed}) to the logical predicates present in the target grammar (e.g., \textit{is$\_$healed}).
Lexicons are good at modeling the compositionality of logic \cite{kamath2019SurveySemanticParsing}.
However, building such a lexicon requires domain expertise and is usually a time-consuming process.
Also, while rule-based systems may perform well on a specific dataset for which they are implemented, they often fail to generalize across a broader variety of questions beyond what is present in the original dataset.
Neural semantic parsers overcome some of these shortcomings by diminishing the need to use a lexicon.
This, in turn, also improves the generalizability of the systems, i.e., their capability to accommodate a wider variety of data.
For instance, given a word with its corresponding entry present in the lexicon (e.g., \textit{healed} $\rightarrow$ \textit{is$\_$healed}), a lexicon-based system may fail to understand a different synonym of the word (in clinical context) that is unavailable in the lexicon (e.g., \textit{repaired}, \textit{mended}).
Thus, the reach of such systems (based on a lexicon) are somewhat limited to the phrases available in their lexicon.
The neural systems use advanced embeddings for representing the words in a query that plays an important role in identifying synonyms and even misspellings \cite{kalyan2020SECNLPSurveyEmbeddings}.

Another requirement of a traditional semantic parser is feature-engineering for its ML component.
It is a crucial step in building ML models, where input features are extracted from raw data \cite{zheng2018FeatureEngineeringMachine}.
It requires domain knowledge and is usually a time-consuming process along with being error-prone due to human involvement.
More importantly, since feature selection and extraction are based on a given dataset, these techniques are oftentimes not generalizable to other kinds of datasets.
Neural SP approaches alleviate the need to build manual features by using advanced forms of representations (such as word and character embeddings) to pass input text into the model \cite{kamath2019SurveySemanticParsing}.

Despite the numerous advantages to using the neural semantic parsers, a traditional system with a well-engineered manually-built lexicon can still outperform them on a given dataset, if not generalize well.
However, especially given the ease of setting up a neural SP system and its advantages, analyzing the errors made by neural semantic parsers is a significant step toward improving these systems in a clinical context.
Thus, in this paper, we systematically assess the efficacy of neural SP when applied to the task of EHR QA.
To identify the common errors made by such models, we use two different neural SP models in our evaluations.
Also, we evaluate these models on two different clinical SP datasets to identify the usual types of errors seen in the clinical context.
We also report the results of traditional SP techniques on all the evaluated datasets for a rounded evaluation.
Our error analysis surfaces the most frequent types of errors made by neural models on clinical datasets that, along with our detailed discussion to tackle these types of errors, can serve as a starting point for future research in this domain.
To our knowledge, this is the first work to systematically assess the performance of neural SP techniques for EHR QA.

\section{Background}
\label{background}

\subsection{Clinical Natural Language Understanding}
\label{clinicalqa}

\vspace{-3pt}
\paragraph{Data:}

Much work is done for building datasets to understand the information need from the clinical questions, specifically directed toward the EHRs \cite{patrick2012OntologyClinicalQuestions, roberts2015NaturalLanguageInterface, roberts2016AnnotatingLogicalForms, pampari2018EmrQALargeCorpus, soni2019UsingFHIRConstruct}.
Pampari et al. \cite{pampari2018EmrQALargeCorpus} take a template-based approach to building a large corpus of question-LF pairs (based on a different representation for LFs) exploiting an existing set of natural language processing (NLP) annotations from the i2b2 datasets.
Though this dataset, named emrQA, is large (with around 1 million question-LF pairs), the variety of questions is limited due to templatization (as is also found in a separate systematic analysis of emrQA's machine comprehension data \cite{yue2020ClinicalReadingComprehension}) and thus this dataset is not representative of the real world clinical QA.
\textcolor{black}{
Likewise, Wang et al. \cite{wang2020TexttoSQLGenerationQuestion} built a large dataset of question-SQL (structured query language) pairs using SQL-based templates where they first automatically generated a set of questions and later filtered/rephrased them through crowd-sourcing.
Again, the use of templates restricts the variety of questions in this dataset to only those that have a SQL query with the same underlying structure as one of the templates.
}
Patrick \& Li \cite{patrick2012OntologyClinicalQuestions} built a dataset of clinical questions collected from the physicians and staff at an intensive care unit (ICU) setting.
Later, Roberts \& Demner-Fushman \cite{roberts2015NaturalLanguageInterface, roberts2016AnnotatingLogicalForms} designed a grammar based on standard $\lambda$-calculus to represent the meaning of clinical questions and annotated the aforementioned set of questions with their corresponding LFs.
Another work used the same $\lambda$-calculus grammar to build a dataset of clinical questions annotated with their LFs in a study using Fast Healthcare Interoperability Resources (FHIR) \cite{healthlevelseveninternationalWelcomeFHIR}.
We use these two clinical question-LF datasets for our analysis as the underlying $\lambda$-calculus grammar allows for a rich representation of the information need from questions and also because several general-domain neural SP models have been developed for this kind of grammar.

\vspace{-15pt}
\paragraph{Methods:}

Another strand of work in this domain focuses on methods for clinical natural language understanding \cite{patrick2012OntologyClinicalQuestions, roberts2017SemanticParsingMethod, pampari2018EmrQALargeCorpus, ruan2019QAnalysisQuestionanswerDriven, schwertner2019FosteringNaturalLanguage}.
Patrick \& Li \cite{patrick2012OntologyClinicalQuestions} built a multilayer classification model to classify the clinical questions into their corresponding templates using rules and traditional ML.
Such a technique is severely limited to the set of templates used for building a model and does not capture the whole spectrum of information representation achievable through SP.
Another work by Ruan et al. \cite{ruan2019QAnalysisQuestionanswerDriven} leveraged knowledge graphs to build a SP system specifically focused on statistical clinical queries in Chinese.
But, their tool supports few statistical operations and the vocabulary is limited.
Schwertner et al. \cite{schwertner2019FosteringNaturalLanguage} built a Portuguese QA system using knowledge bases where the semantic understanding engine was based on an external system.
Conversely, two studies \cite{neuraz2018NaturalLanguageUnderstanding, neuraz2020ImpactSpecializedCorpora} tackled the task of shallow SP (for a dialogue system) where the main focus is on filling slots of information from a given utterance (not on mapping it to a meaning representation).
In one, Neuraz et al. \cite{neuraz2018NaturalLanguageUnderstanding} evaluated the efficacy of expanding their existing dataset using an external machine translation system (Google Translate) while in another, Neuraz et al. \cite{neuraz2020ImpactSpecializedCorpora} experimented with different word embeddings trained on general and biomedical domain data.
Pampari et al. \cite{pampari2018EmrQALargeCorpus} applied a baseline neural-based model on the emrQA dataset but did not conduct a thorough analysis of the errors made by neural models.
Similarly, Wang et al. \cite{wang2020TexttoSQLGenerationQuestion} proposed a neural model based on sequence-to-sequence networks with attention-based copying.
They, however, do not assess the efficacy of their technique in comparison to the rule-based alternatives.
Roberts \& Patra \cite{roberts2017SemanticParsingMethod} developed a SP pipeline using rules (to generate a candidate set of LFs) and traditional ML (to select the best candidate thereafter).
In this work, we use this SP methodology as the lexicon-based system and compare the performance against the neural alternatives.

\vspace{-10pt}
\subsection{General-domain Neural Semantic Parsing}
\label{neuralsp}
\vspace{-5pt}

Neural SP has had much success in the general domain \cite{kamath2019SurveySemanticParsing}, particularly using the sequence-to-sequence (Seq2Seq) models \cite{sutskever2014SequenceSequenceLearning}.
Such methods for parsing input queries to their corresponding LFs can be broadly divided into $3$ categories on the basis of intermediate and target meaning representations (MRs) used for training.

\paragraph{Direct Learning:}

The models in first category \textit{directly} learns to map input queries to their LFs in target grammar.
Here, the approaches do not explicitly define an intermediate meaning representation (such models do however use an implicit IR that is usually not insightful).
E.g., Dong \& Lapata \cite{dong2016LanguageLogicalForm} tackle the SP task using an encoder-decoder scheme with attention where they pass each sequence of the query as input to the encoder and generate a LF using the decoder.
They proposed a decoder that takes into account the hierarchical structure of the LFs.
To overcome a common problem with these methods related to rare words, a series of works came about anonymizing the entities with corresponding types (using an ontology).
E.g., mapping \textit{Paracetamol} to its type \textit{Medicine}.
Specifically, Jia \& Liang \cite{jia2016DataRecombinationNeural} proposed a method using attention-based copying where, alongside the normal decoding mechanism of employing softmax over the entire vocabulary, the decoder can choose to copy words directly from the input query.

\vspace{-15pt}
\paragraph{Using Intermediate MR:}

The second category of models employs abstract representations as intermediate forms while parsing the input to its respective LF.
This type of abstraction allows sharing between examples and helps models to better generalize on smaller datasets \cite{kamath2019SurveySemanticParsing}.
E.g., Cheng et al. \cite{cheng2017LearningStructuredNatural} build insightful intermediate representations (in natural language) using a transition system and map them to target representations thereafter.
Similarly, Dong \& Lapata \cite{dong2018CoarsetofineDecodingNeural} develop a \textsc{Coarse2Fine} model that generates a coarse intermediate representation (where granular details of the LFs are masked) that is passed through the decoder to predict the final LF (with all the fine details).

\vspace{-15pt}
\paragraph{Constrained Grammar:}

The last category of models use a constrained grammar that controls the decoder while generating the derivations for LFs.
E.g., Krishnamurthy et al. \cite{krishnamurthy2017NeuralSemanticParsing} use a type-constraining grammar to guide the decoder.
In other words, the grammar ensures that the decoder predicts LFs satisfying the type constraints.
Similarly, the \textsc{Tranx} model by Yin \& Neubig \cite{yin2018TRANXTransitionbasedNeural} employs abstract syntax description language (ASDL) framework to learn a general-purpose representation in the form of abstract syntax tree (AST).
The ASTs are not passed through a decoder, instead, they are mapped to the final LFs using a deterministic user-defined function.
Such an architecture releases the model from learning the target grammar from inadequate training data.

Looking at the advantages offered by the last two categories for smaller datasets (such as the clinical datasets), we choose to evaluate one model from each.
Specifically, we use the \textsc{Coarse2Fine}  (by Dong \& Lapata \cite{dong2018CoarsetofineDecodingNeural}) and \textsc{Tranx} (by Yin \& Neubig \cite{yin2018TRANXTransitionbasedNeural}) 
models for our analysis.
We also choose to modify an existing direct learning model (Jia \& Liang \cite{jia2016DataRecombinationNeural}) by incorporating transformers \cite{vaswani2017attention} and recombination techniques \cite{jia2016DataRecombinationNeural} to augment our data.

\section{Materials and Methods}

\subsection{Data}
\label{data}
\vspace{-5pt}

We use two clinical QA datasets for our evaluations.
Both the datasets consist of patient-specific clinical questions (that can be answered using EHR data) and their corresponding LFs based on $\lambda$-calculus.

\vspace{-15pt}

\paragraph{\textsc{ICU\textsubscript{data}}:}

Questions in one dataset were originally collected from the staff at an intensive care unit (ICU) setting through interviews and observing the clinical workflow \cite{patrick2012OntologyClinicalQuestions}.
This set of questions was further annotated with their meaning representations in a separate study \cite{roberts2016AnnotatingLogicalForms}.
We use a set of $401$ deduplicated questions from this study.

\vspace{-15pt}

\paragraph{\textsc{FHIR\textsubscript{data}}:}

Another dataset was constructed in a study that demonstrates the efficacy of an annotation tool based on Fast Healthcare Interoperability Resources (FHIR) \cite{healthlevelseveninternationalWelcomeFHIR} to reduce the overall LF annotation time by automating the time-consuming step of concept normalization annotation and better aligning the normalized concepts to a specific EHR implementation\cite{soni2019UsingFHIRConstruct}.
The questions were created (and annotated with their LFs) by a physician and a biomedical informatics doctoral student as they viewed EHR data using the constructed FHIR tool.
This work constructed a corpus of $1000$ questions annotated with their meaning representations using the same $\lambda$-calculus grammar as \textsc{ICU\textsubscript{data}} \cite{roberts2016AnnotatingLogicalForms}.
We include a deduplicated set of $980$ questions in our analyses.

\vspace{-15pt}

\paragraph{Logical Forms}
The LFs in all these datasets consist of two main elements, namely, predicates and parameters, along with the quantifier $\lambda$.
Predicates are functions that are used to retrieve and manipulate event information.
For instance, consider the following question (Q) and LF pair from the \textsc{ICU\textsubscript{data}}.

\begin{center}
\vspace{-12pt}
\textbf{Q:} What microorganisms were have been grown? \phantom{xxxx}
\textbf{LF:} \textit{$\lambda x$.has\_concept($x$, C2242979, visit)}
\vspace{-12pt}
\end{center}

\begin{table*}[t]
\caption{Example Qs. \textbf{Q} -- Question. \textbf{LF} -- Logical Form. \textbf{ES} -- Post entity substitution. \textbf{PP} -- Post preprocessing.} \label{tab:data-example}
\vspace{-7pt}
\centering
\small
\begin{tabular}{c l}
\hline 
\textbf{Dataset} & \textbf{Example} \\
\hline
\multirow{4}{*}{\textbf{\textsc{ICU\textsubscript{data}}}} & \textbf{Q}: \space Did \textit{her} \textit{temperature} fall below \textit{38C}? \\

& \textbf{LF}: \textit{delta($\lambda x$.has\_concept($x$, cui) $\wedge$ less\_than($x$, `38C'))} \\

& \textbf{ES}: Did \textit{patient} \textit{concept(C0005903)} fall below \textit{measurement(`38C')} ? \\
		
& \textbf{PP}: did patient concept fall below measur \\ 
\hline 
\multirow{4}{*}{\textbf{\textsc{FHIR\textsubscript{data}}}} & \textbf{Q}: \space How many times were the \textit{influenza shots} given to \textit{her} \textit{in the past 3 years}? \\

& \textbf{LF}: \textit{count($\lambda x$.has\_concept($x$, cui) $\wedge$ time\_within($x$, `38C'))} \\

& \textbf{ES}: How many times were the \textit{concept(C0234422)} given to \textit{patient} \textit{temporal\_ref(`in the past 3 years')} ? \\
		
& \textbf{PP}: how mani time were the concept given to patient temporal\_ref \\ 
\hline 
\end{tabular}
\end{table*}

Here, \textit{has$\_$concept} is a predicate in the clinical grammar for retrieving all the concepts with a given \textit{cui} (a unique concept identifier) and implicit time frame (temporal information to use while retrieving the concepts from the EHRs) as parameters.
The \textit{cui} is usually normalized to some standard vocabulary (such as the UMLS \cite{lindberg1993UnifiedMedicalLanguage}).
In this case, \textit{C2242979} is a UMLS code for \textit{Microbial culture}.
The other parameter \textit{visit} is the implicit time frame that restricts the retrieved concepts to the current inpatient/outpatient visit.

Other predicates are frequently applied with the aforementioned \textit{has$\_$concept} predicate to express different information needs of the natural language questions.
Consider the following example.

\begin{center}
\vspace{-12pt}
    \textbf{Q:} What is the volume of his urine last night? \\
    \textbf{LF:} \textit{sum($\lambda x$.has\_concept($x$, C0042036, visit) $\wedge$ time\_within($x$, `last night'))}
\vspace{-12pt}
\end{center}

In this instance, the \textit{time$\_$within} predicate, when applied in conjunction with the \textit{has$\_$concept}, further filters down the retrieved concepts using the explicit temporal reference from the question, i.e., \textit{``last night''}.
Further, the \textit{sum} predicate adds up the values of the given set of concepts and return that information.
Some other examples include \textit{latest} (selects the most recent concept from a given set of concepts), \textit{is$\_$negative} (returns whether a given concept is positive or negative in the clinical context), \textit{count} (returns the total number of concepts in a given set), and \textit{reason} (returns the reason associated with a given concept).
Though all the predicates in a LF have some direct or indirect relation to the natural language question, the outermost predicate can provide a good sense of the question and answer types.
E.g., the outermost predicates for the aforementioned examples are \textit{has$\_$concept} and \textit{sum}.

\vspace{-10pt}
\subsection{Preprocessing}
\vspace{-5pt}

Most of the neural SP models assume the entities in input queries to be identified \cite{dong2018CoarsetofineDecodingNeural, yin2018TRANXTransitionbasedNeural, kamath2019SurveySemanticParsing}.
We thus follow suit and replace the entities in the clinical datasets such as person references, temporal references, and measurements by identifiers \textit{person}, \textit{temporal\_ref}, and \textit{measurement}.
The clinical entities in the questions are replaced by \textit{concept}.
This abstraction step ensures a proper evaluation of the semantic parser without error propagation from other tasks such as entity extraction.
We exclude the implicit time frames to be consistent with the evaluated models' LF format.

Further, we apply some textual preprocessing steps to the clinical datasets in order to conform to the input and output data requirements of the neural models used in our evaluations.
Specifically, for all datasets we (i) lowercase the questions, (ii) apply the Porter stemming algorithm using NLTK, and (iii) remove all punctuation.
An example from each dataset
is shown in Table \ref{tab:data-example}.

Descriptive statistics for all the datasets used in our evaluation are presented in Table \ref{tab:descriptive-stats}.
Note that the \textsc{FHIR\textsubscript{data}} consists of a wider variety of natural language tokens than the \textsc{ICU\textsubscript{data}} ($191$ versus $157$).

\begin{SCtable}
    \small
	\caption{Descriptive statistics.
    All the statistics are on full datasets.
    The statistics are calculated after preprocessing.
    Note that the set of unique predicates for \textsc{FHIR\textsubscript{data}} is a proper subset of the same in \textsc{ICU\textsubscript{data}}.
	}
	\label{tab:descriptive-stats}
	\centering
	\begin{tabular}{c c c}
		\hline 
		\textbf{Metric} & \textbf{\textsc{ICU\textsubscript{data}}} & \textbf{\textsc{FHIR\textsubscript{data}}} \\
		\hline 
		\# of queries & $401$ & $980$ \\
		\# unique tokens & $157$ & $191$ \\
		\# unique predicates & $53$ & $21$ \\
		Avg \# of tokens / query & $5.18$ & $5.84$ \\
		Avg \# of preds / query & $3.64$ & $3.73$ \\
		\hline
	\end{tabular}
\vspace{-25pt}
\end{SCtable}

\vspace{-10pt}
\subsection{Models}
\label{models}
\vspace{-5pt}

We use two neural SP models based on different architectures that are shown to be effective for small datasets.
To view the performance of neural systems in light of a traditional lexicon-based approach, we also report the results of two traditional rule-based and ML approaches for all the evaluated datasets.
We collectively refer to these traditional methods as \textsc{Lexicon-based} as both of them employ some kind of lexicon to map the natural language phrases from the input utterances to logical predicates present in the $\lambda$-calculus.

\vspace{-12pt}

\subsubsection{\textsc{Tranx}}
\vspace{-5pt}

A high-level architecture of this system is presented in Figure \ref{fig:tranx}.
The main component of this model is a transition system that converts a given query to its corresponding abstract syntax tree (AST), used as a general-purpose meaning representation \cite{yin2018TRANXTransitionbasedNeural}.
The transition system employs a domain-specific grammar, defined under the abstract syntax description language (ASDL) framework by the user, to generate ASTs.
After a candidate AST is selected by the neural network (an encoder-decoder framework), it is converted to a domain-specific meaning representation (in our case, this is $\lambda$-calculus) in a deterministic manner.
We make minimal modifications to theexisting ASDL grammar \cite{rabinovich2017AbstractSyntaxNetworks} to accommodate both the clinical datasets.
Specifically, we add a few constructors to the grammar to cover the clinical domain-specific constructs such as \emph{Dose} and \emph{Reason} (that are corresponding to certain logical predicates present in the clinical lexicons such as \textit{dose} and \textit{reason}).
We further define additional mappings in the grammar for predicates such as \emph{has\_concept} and \emph{time\_within}.

\vspace{-12pt}

\subsubsection{\textsc{Coarse2Fine}}
\vspace{-5pt}

The workflow of this system is shown in Figure \ref{fig:coarse2fine}.
The approach for this model is broadly divided into two parts where a series of encoders and decoders are employed to construct intermediate representations \cite{dong2018CoarsetofineDecodingNeural}.
In the earlier part, the input is encoded (using input encoder) into a representation that is used to predict a sketch sequence (using sketch decoder).
A meaning sketch generated by this model is an abstraction over the LFs where more specific (or \textit{fine}) details related to a LF such as variable names and arguments are omitted.
Thus, sketch is a more \textit{coarse} representation of the LFs and are somewhat easier to predict (because of fewer details).
In the later phase, they use this sketch (through sketch encoder) and the input query representation to predict a LF (using sketch-guided output decoder).
The decoders in both the phases use an attention mechanism to better model the alignments.

\vspace{-12pt}

\subsubsection{\textsc{Transformer}}
\vspace{-5pt}

\begin{figure*}[t]
 \centering
 \begin{subfigure}{\textwidth}
     \centering
     \includegraphics[width=0.70\textwidth]{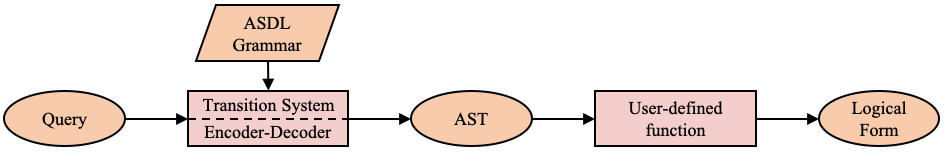}
     \vspace{-5pt}
     \caption{\textsc{Tranx}}
     \label{fig:tranx}
 \end{subfigure}
 \begin{subfigure}{\textwidth}
     \vspace{14pt}
     \centering
     \includegraphics[width=0.70\textwidth]{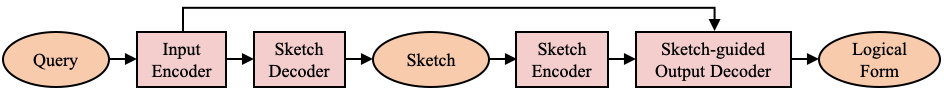}
     \vspace{-5pt}
     \caption{\textsc{Coarse2Fine}}
     \label{fig:coarse2fine}
 \end{subfigure}
 \begin{subfigure}{\textwidth}
     \vspace{15pt}
     \centering
     \includegraphics[width=0.5\textwidth]{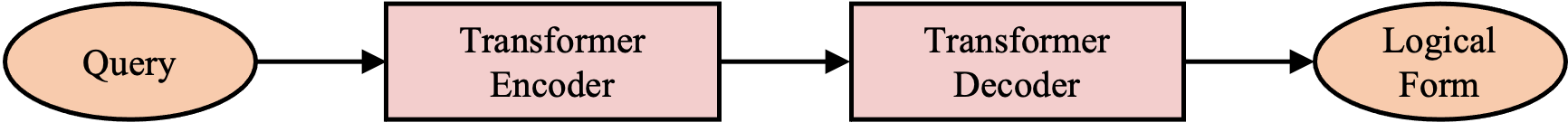}
     \vspace{-5pt}
     \caption{\textsc{Transformer}}
     \label{fig:transformer}
 \end{subfigure}
 \begin{subfigure}{\textwidth}
     \vspace{10pt}
     \centering
     \includegraphics[width=0.70\textwidth]{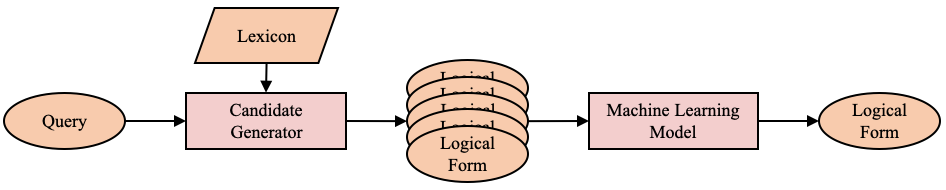}
     \vspace{-5pt}
     \caption{\textsc{Lexicon-based}}
     \label{fig:lexiconbased}
 \end{subfigure}
\vspace{5pt}
\caption{High-level architectures of the evaluated models.}
\label{fig:modelarchitectures}
\end{figure*}

We also employ transformer models \cite{vaswani2017attention} to our SP task (Figure \ref{fig:transformer}).
The model architecture is inspired from Jia \& Liang \cite{jia2016DataRecombinationNeural}, where they used recurrent neural networks (RNNs) for encoding and decoding.
Differently, we explore transformers \cite{vaswani2017attention} for both encoder and decoder.
While RNNs model sequential information in text very well, transformer models are shown to learn bidirectional representations that takes into account textual dependencies from both directions.
One caveat with these powerful models is that they require large amounts of data in order to produce good results.
To overcome this, we also apply data recombination techniques \cite{jia2016DataRecombinationNeural} to expand the size of our datasets.
Specifically, we use $3$ types of augmentations, namely, entity abstraction (replace entities in questions with other similar entities to create new questions), whole phrase abstraction (replace key phrases in the dataset questions that represents similar meaning), and concatenation (combine different questions to create larger harder queries for the model).

\vspace{-12pt}

\subsubsection{\textsc{Lexicon-based}}
\vspace{-5pt}

An overall architecture of the models used in our evaluations is given in Figure \ref{fig:lexiconbased}.
Both the systems use some kind of candidate generator to produce a set of candidate LFs (or representations that can be trivially converted to LFs).
Then, an ML model is used to rank the candidates from which the top ranked item is returned as the predicted LF.
The method used for the clinical datasets employs a rule-based pipeline to generate a set of candidate LFs \cite{roberts2017SemanticParsingMethod}.
Then, a support vector machine (SVM) is employed to select the best LF based on the features extracted from the input utterance and the generated candidates.

\vspace{-10pt}

\subsection{Evaluation}
\vspace{-5pt}

The models are trained on a subset of the dataset called a \textit{train} set while evaluating their performances during the training on a \textit{development} set for early stopping.
Finally, the selected models (after early stopping) are evaluated on a \textit{test} set to get the reported performances.
\textcolor{black}{
We report the \textit{accuracy} of the SP models that measures the fraction of input queries for which the predicted LF exactly matches the gold annotated LF.
}

\vspace{-12pt}

\subsubsection{Cross-validation}
\vspace{-5pt}

For both the clinical datasets, the train, dev, and test splits are not well-defined.
Thus, for the neural models, we use a $10$-fold cross-validation (CV) scheme to evaluate the performance.
Specifically, we first split each clinical dataset into $10$ equal (or nearly equal) parts or folds.
Then, each fold is treated as a test set while the model is trained on the other $8$ folds keeping $1$ fold as a development set.
We report the average performance of the models after considering each fold as a test set exactly once, i.e., running the model $10$ times.
In the case of \textsc{Lexicon-based} model, we report the results using a leave-one-out CV (LOOV) evaluation scheme.
LOOV is a special (more extensive) type of CV where the number of folds is equal to the total number of instances in a dataset.
Precisely, each instance in the dataset is kept aside for testing exactly once while the model is trained on all the remaining instances.

\vspace{-12pt}

\subsubsection{Cross-dataset}
\vspace{-5pt}

Alongside the CV evaluation for the clinical datasets, we also experiment with training the neural models in a cross-dataset (CD) fashion.
Here, a model is trained on one clinical dataset and then its performance is evaluated on another clinical dataset.
Particularly, we use the best performing models from our CV evaluation for each dataset and apply them to a different clinical dataset, considering the whole target dataset as a \textit{test} set.

\vspace{-10pt}
\subsection{Experimental setting}
\vspace{-5pt}

We use the recommended set of hyperparameters from the \textsc{Tranx} and \textsc{Coarse2Fine} papers.
For \textsc{Tranx}, the hidden vectors size is $256$; the word vector dimension is $128$; the learning rate is $0.0025$; the maximum number of epochs is $100$ with early stopping using a training patience of $5$.
For \textsc{Coarse2Fine}, the hidden vectors size is $300$; the word vector dimension is $150$; the learning rate is $0.005$; the max number of epochs is $100$ validating every $10$ epochs for early stopping.
For both the models, the dropout rate is $ 0.5 $ and the learning rate decay is $0.985$.
The word embeddings are initialized using GloVe \cite{pennington2014GloveGlobalVectors}.
For the \transformer\ model, the number of examples added using recombination is $1800$ for train and $300$ for dev; the number of epochs is $200$; learning rate is $5e$-$4$.

\section{Results}
\label{results}
\vspace{-5pt}

The performance of the evaluated models is presented in Table \ref{tab:allmodelscv}.
In both CV and CD settings, the \lexicon\ model performed the best as compared to the other neural alternatives, highlighting the importance of domain-specific ``manual'' fine-tuning that goes into constructing these.
In CV setting, the \textsc{Coarse2Fine} model performed consistently better than both the other neural models.
While the performance of all the models on the \textsc{ICU\textsubscript{data}} is promising, their accuracies on the \textsc{FHIR\textsubscript{data}} is much higher (about $15$ points better for the neural models).

In CD setting, the performance of all the models on the target datasets dropped after training on a different clinical dataset.
Note that the reduction in accuracy for the \textsc{FHIR\textsubscript{data}} (more than $25$ points) is larger than that for the \textsc{ICU\textsubscript{data}} (around $4$-$7$ points).
Contrary to CV results, all the models performed better on the \textsc{ICU\textsubscript{data}} than on the \textsc{FHIR\textsubscript{data}} and the \transformer\ model achieved the best score on \textsc{ICU\textsubscript{data}} while the \lexicon\ model performed the best on \textsc{FHIR\textsubscript{data}}.
Interestingly, the \transformer\ model did not hold good performance during the cross-data evaluation on \fhirdata.
Also, \lexicon\ models suffered the worst during CD on \icudata.

The errors made by all the models for a given dataset and evaluation method may be representative of the typical errors made by such models in these configurations (dataset and evaluation).
Thus, we report the proportion of errors made by both the neural models for a given corpus and evaluation method in Table \ref{tab:shortcomings}.
The example on the first row shows a question where the \lexicon\ model correctly predicted the LF but all the other models failed.
Here, the reason may be domain-specific artifacts of data (such as \textit{fistula closed} being translated to \textit{is\_healed} predicate) that are incorporated into the lexicon built for these models.
On the other hand, referring to the second example under \fhirdata\ (asking about \textit{MMR} vaccine start date), the complex syntactic information in the question was not properly parsed by the \lexicon\ system.
We further analyze these errors and present some of the prominent types of errors made by the MR models in Table \ref{tab:erroranalysis}.
We note that the most wrongly predicted answer types for both the datasets include \textit{is$\_*$} and \textit{$\lambda x$} followed by \textit{sum} for \textsc{ICU\textsubscript{data}} and \textit{reason} for \textsc{FHIR\textsubscript{data}}.
Interestingly, among the few instances of \textit{count} in the \textsc{ICU\textsubscript{data}}, both the models failed to predict correctly in the case of CV while in a cross-dataset evaluation not all the models made such an error.
Similarly, the LFs starting with \textit{earliest} in the \textsc{FHIR\textsubscript{data}} were all incorrectly predicted by both the models in the case of cross-dataset evaluation while it was not the case in CV.

\begin{table}
\small
    \caption{Cross-validation and cross-dataset evaluation.} \label{tab:allmodelscv}
    \vspace{-7pt}
	\centering
	\begin{tabular}{c c c c c}
		\hline 
		\multirow{2}{*}{\textbf{Model}} & \multicolumn{2}{c}{\textbf{Cross-validation}} & \multicolumn{2}{c}{\textbf{Cross-dataset}}\\
		\cmidrule(r){2-3} \cmidrule(l){4-5}
		& \textbf{\textsc{ICU\textsubscript{data}}} & \textbf{\textsc{FHIR\textsubscript{data}}} & \textbf{\textsc{ICU\textsubscript{data}}} & \textbf{\textsc{FHIR\textsubscript{data}}} \\ 
		\hline
		\textsc{Tranx} & $77.5$ & $93.1$ & $71.8$ & $67.7$ \\
		\hline
		\textsc{Coarse2Fine} & $80.5$ & $94.7$ & $72.8$ & $66.2$ \\ 
		\hline
		\textsc{Transformers} & $79.3$ & $93.9$ & $74.8$ & $45.9$ \\
		\hline 
		\textsc{Lexicon-based} & $86.8$ & $97.6$ & $74.6$ & $71.7$ \\
		\hline
	\end{tabular}
\end{table}

\begin{table*}[t]
    \vspace{5pt}
    \small
	\caption{Error analysis. \textbf{MR} -- models using intermediate machine representations (\tranx\ and \cf). \textbf{TF} -- \transformer. \textbf{LB} -- \lexicon. \textbf{Q} -- Question. \textbf{L} -- Simplified logical form for brevity.} \label{tab:shortcomings}
	\vspace{-7pt}
	\centering
	\begin{tabular}{cccclc}
		\hline 
		\multirow{2}{*}{\textbf{Corpus}} & \multicolumn{3}{c}{\textbf{Model Prediction}} & \multicolumn{1}{c}{\multirow{2}{*}{\textbf{Example}}} & \multirow{2}{*}{\textbf{Count}} \\ 
		\cline{2-4}
		 & \textbf{MR} & \textbf{TF} & \textbf{LB} & & \\
		\hline
		\multirow{9}{*}{\textsc{ICU\textsubscript{data}}}
			  & \xmark & \xmark & \cmark
			 & \begin{tabular}{@{}l@{}}\textbf{Q:} Has the fistula closed? \\ \textbf{L:} \textit{delta($\lambda x$.has\_concept($x$) $\wedge$ is\_healed($x$))}\end{tabular}
			 & 18 \\
		\cline{2-6}
		& \cmark & \cmark & \xmark
			 & \begin{tabular}{@{}l@{}}\textbf{Q:} How much sedation is she on? \\ \textbf{L:} \textit{latest($\lambda x$.has\_concept($x$))}\end{tabular}
			 & 7 \\
		\cline{2-6}
		& \cmark & \xmark & \xmark
			 & \begin{tabular}{@{}l@{}}\textbf{Q:} How did they confirm H1N1? \\ \textbf{L:} \textit{latest($\lambda x$.has\_concept($x$)))}\end{tabular}
			 & 2 \\
		\cline{2-6}
		& \xmark & \cmark & \xmark
			 & \begin{tabular}{@{}l@{}}\textbf{Q:} How was the neurological state overnight? \\ \textbf{L:} \textit{$\lambda x$.has\_concept($x$) $\wedge$ time\_within($x$)}\end{tabular}
			 & 5 \\
	    \cline{2-6}
		& \xmark & \xmark & \xmark
			 & \begin{tabular}{@{}l@{}}\textbf{Q:} Has his culture grown any microorganisms? \\ \textbf{L:} \textit{is\_positive(latest($\lambda x$.has\_concept($x$))))}\end{tabular}
			 & 32 \\
		\hline 
        \multirow{9}{*}{\textsc{FHIR\textsubscript{data}}}
			  & \xmark & \xmark & \cmark
			 & \begin{tabular}{@{}l@{}}
			 		\textbf{Q:} What part had laceration? \\ 
			 		\textbf{L:} \textit{location(latest($\lambda x$.has\_concept($x$))))}
			 	\end{tabular}
			 & 17 \\
		\cline{2-6}
		& \cmark & \cmark & \xmark
			 & \begin{tabular}{@{}l@{}}
			 		\textbf{Q:} When was the MMR started? \\ 
			 		\textbf{L:} \textit{time(earliest($\lambda x$.has\_concept($x$))))}
			 	\end{tabular}
			 & 8 \\
		\cline{2-6}
		& \cmark & \xmark & \xmark
			 & \begin{tabular}{@{}l@{}}
			 		\textbf{Q:} How much did he weigh in the last measurement? \\ 
			 		\textbf{L:} \textit{latest($\lambda x$.has\_concept($x$)))}
			 	\end{tabular}
			 & 3 \\
		\cline{2-6}
		& \xmark & \cmark & \xmark
			 & \begin{tabular}{@{}l@{}}
			 		-- \\ 
			 		--
			 	\end{tabular}
			 & 0 \\
	    \cline{2-6}
		& \xmark & \xmark & \xmark
			 & \begin{tabular}{@{}l@{}}
			 		\textbf{Q:} What about her animal dander allergy? \\ 
			 		\textbf{L:} \textit{$\lambda x$.has\_concept($x$)}
			 	\end{tabular}
			 & 8 \\
		\hline
	\end{tabular}
\end{table*}

\begin{table*}[t]
    \small
	\caption{Incorrect predictions made by both \tranx\ and \cf\ models in case of different evaluation strategies. Variant refers to the questions with logical form having the given predicate/quantifier at the outermost position. \textbf{CV} -- Cross-validation. \textbf{CD} -- Cross-dataset. \textbf{Q} -- Question. \textbf{L} -- Simplified logical form excluding some parameters for brevity.} \label{tab:erroranalysis}
	\vspace{-7pt}
	\centering
	\begin{tabular}{cclcccc}
		\hline 
		\multirow{2}{*}{\textbf{Corpus}} & \multicolumn{3}{c}{\textbf{Variant}} & $\!\!\!$ & \multicolumn{2}{c}{\textbf{Errors}} \\
		\cline{2-4} \cline{6-7}
		 & \textbf{Variant} & \multicolumn{1}{c}{\textbf{Example}} & \textbf{Count} &$\!\!\!$& \textbf{CV} & \textbf{CD} \\ 
		\hline
		\multirow{7}{*}{\textsc{ICU\textsubscript{data}}} & \textit{is$\_*$} & \begin{tabular}{@{}l@{}}\textbf{Q:} Was the blood culture negative? \\ \textbf{L:} \textit{is\_negative(latest($\lambda x$.has\_concept($x$)))}\end{tabular} & 20 & $\!\!\!$ & $11$ & $19 $ \\
		\cline{2-7}
		                                            & \textit{$\lambda x.*$} & \begin{tabular}{@{}l@{}}\textbf{Q:} What were the drain outputs for the past 12 hours? \\ \textbf{L:} \textit{$\lambda x$.has\_concept($x$) $\wedge$ time\_within($x$)}\end{tabular} & 19 & $\!\!\!$ & $10$ & $16$ \\
		\cline{2-7}
		                                            & \textit{sum}           & \begin{tabular}{@{}l@{}}\textbf{Q:} What is the volume of his urine last night? \\ \textbf{L:} \textit{sum($\lambda x$.has\_concept($x$) $\wedge$ time\_within($x$))}\end{tabular} & 33 & $\!\!\!$ & $5$ & $all$ \\
		\cline{2-7}
		                                            & \textit{count}         & \begin{tabular}{@{}l@{}}\textbf{Q:} How often were they bleeding? \\ \textbf{L:} \textit{count($\lambda x$.has\_concept($x$))}\end{tabular} & 3 & $\!\!\!$ & $all$ & $none$ \\
		\hline 
		\multirow{7}{*}{\textsc{FHIR\textsubscript{data}}} & \textit{is\_*}  & \begin{tabular}{@{}l@{}}\textbf{Q:} Are his triglycerides getting low? \\ \textbf{L:} \textit{is\_decreasing($\lambda x$.has\_concept($x$))}\end{tabular} & 28 & $\!\!\!$ & $5$ & $27$ \\
		\cline{2-7}
		                                            & \textit{$\lambda x.*$} & \begin{tabular}{@{}l@{}}\textbf{Q:} What is her vaccination history for polio? \\ \textbf{L:} \textit{$\lambda x$.has\_concept($x$)}\end{tabular} & 9 & $\!\!\!$ & $7$ & $4$ \\
		\cline{2-7}
		                                            & \textit{reason}        & \begin{tabular}{@{}l@{}}\textbf{Q:} What was the recent office visit for? \\ \textbf{L:} \textit{reason(latest($\lambda x$.has\_concept($x$)))}\end{tabular} & 29 & $\!\!\!$ & $2$ & $all$ \\
		\cline{2-7}
		                                            & \textit{earliest}      & \begin{tabular}{@{}l@{}}\textbf{Q:} What are the details of her first allergy screening? \\ \textbf{L:} \textit{earliest($\lambda x$.has\_concept($x$))}\end{tabular} & 5 & $\!\!\!$ & $none$ & $all$ \\
		\hline
	\end{tabular}
\end{table*}

\section{Discussion}
\label{discussion}
\vspace{-5pt}

In the case of CV, both the neural models achieve a lower accuracy on the \textsc{ICU\textsubscript{data}} than that on the \textsc{FHIR\textsubscript{data}} (see Table \ref{tab:allmodelscv}).
This can be related to the higher diversity of logical predicates present in the \textsc{ICU\textsubscript{data}}.
Specifically, the \textsc{ICU\textsubscript{data}} contains a total of $53$ unique logical predicates while the \textsc{FHIR\textsubscript{data}} has $21$ distinct predicates (refer Table \ref{tab:descriptive-stats}).
This difference may have led to an improved performance in the case of the \textsc{FHIR\textsubscript{data}}, as the neural models had to generalize on a smaller number of unique predicates.
The models performed better on the \textsc{FHIR\textsubscript{data}} regardless of the fact that it contains a higher number of unique tokens, i.e., it is more diverse in terms of natural language words.
However, the pre-trained word embeddings (GloVe \cite{pennington2014GloveGlobalVectors}) may have helped the models in interpreting the different natural language tokens and thus reduced any performance differences that may have occurred because of this.

On the other hand, the \textsc{Lexicon-based} systems performed with almost the same levels of performance on both the clinical datasets (a little better on the \textsc{ICU\textsubscript{data}}).
The reason behind not seeing much difference in the performance of these traditional systems may be rooted in the rule-based backbone of these methods that uses a hand-built lexicon to generate a set of candidate LFs.
The higher the number of questions with a correct LFs in its candidates (i.e., higher the coverage), the higher the chance is for the final ML model to predict correctly.
Thus, the quality of lexicons impacts the overall performance of such semantic parsers, more so than the variety of logical predicates present in a dataset.
The \textsc{Lexicon-based} methods achieved the best scores for each of the datasets.
Also, the \textsc{Coarse2Fine} model performed consistently better than the \textsc{Tranx} model on both the datasets.

The cross-dataset results maintains the same consistency in terms of the relative performances of the models on the clinical datasets (see Table \ref{tab:allmodelscv}).
However, both the models suffered a performance drop when trained on a different dataset and tested on another.
Such a drop in the performance was also seen in the task of machine comprehension when various transformer-based deep learning models were tested in a cross-dataset fashion (i.e., fine-tuned on one clinical dataset and tested on another) \cite{soni2020EvaluationDatasetSelection}.
This observed performance difference highlights the differences between both the evaluated clinical SP datasets in terms of the variety of logical predicates and natural language tokens.
This, in turn, also suggests the requirement of a generalizable clinical SP dataset.

Some of the most common errors made by both the neural models in each evaluation setting were related to questions with a LF starting with \textit{is$\_*$}.
Since the specific methods of this form have fewer instances in the entire dataset, the models could not figure out the correct natural language terms to be mapped to these predicates.
For most of such cases, the models incorrectly predicted a LF starting with \textit{delta} (a predicate that checks if a given set is non-empty) or in other cases beginning with a different \textit{is$\_*$} predicate.
Note that the answer type for both the variety of predicates (\textit{is$\_*$} and \textit{delta}) is same, i.e., yes or no.
So the models were able to map the questions to an appropriate answer type but failed to interpret the correct information need of a given question in such cases.

Interestingly, for the questions with LF having \textit{$\lambda x$} at the outermost position,
we found that almost all the incorrect predictions resulted in a LF starting with \textit{latest} (a predicate that returns the most recent concept from a given set).
Here, even the answer types were wrongly interpreted by the neural models, in that a LF that returns a single concept is generated instead of the one that returns a set.
We hypothesize that it could be because of the similarity of questions with LFs having \textit{$\lambda x$} or \textit{latest} at the outermost place (refer the examples for \textit{$\lambda x$} and \textit{earliest} in Table \ref{tab:erroranalysis}).

Not surprisingly, the most common type of errors found during the cross-dataset evaluation were related to the presence of certain predicates in the training set.
Consider, for example, the questions with LFs starting with \textit{sum} for the \textsc{ICU\textsubscript{data}} and starting with \textit{reason} or \textit{earliest} for the \textsc{FHIR\textsubscript{data}}.
In all these cases, the models failed to predict even a single instance correctly.
The common thing between these cases is that the predicate at hand was either not present in the training set or only a handful of those were present.
For the examples shown in Table \ref{tab:erroranalysis}, \textit{sum} is not at all present in the \textsc{FHIR\textsubscript{data}} while there is only one question for each of the other variants, \textit{reason} and \textit{earliest}, in the \textsc{ICU\textsubscript{data}}.
We confirmed that the \textsc{ICU\textsubscript{data}} fold used for training the model that was used for the cross-dataset evaluation contained these individual instances for the $2$ aforementioned variants (\textit{reason} and \textit{earliest}).
Also, for the variant \textit{count}, there are only $3$ examples in the \textsc{ICU\textsubscript{data}} to learn from while there are a total of $111$ instances in the \textsc{FHIR\textsubscript{data}}.
Again, interestingly, the models performed poorly in the cross-validation setting while they predicted all the examples correctly in the cross-dataset setting.
Hence, though the neural models can be used without the laboriously built lexicons, the variety and availability of the different kinds of predicates in the training set plays an important role in their overall performance.

For the scope of this study and being consistent with the task of SP as tackled by the evaluated neural models, we took several steps to map our clinical SP task.
Unlike the general domain, it is much harder to extract and normalize the clinical concepts due to many reasons such as limited availability of the annotated datasets (due to concerns such as privacy) and the differences in EHR standards and physician preferences \cite{fu2020ClinicalConceptExtraction}.
Also, the implicit time frames play an important role in interpreting and retrieving the exact information need from a given question.
Though it can be classified separately and added to the resulting LFs from these models, it is important to note that such a classifier may add another layer of error in the process to generating an extended LF.
In the future, we plan to stack all the key components involved in a clinical SP system, to go from raw question to LF, and evaluate the impact of each step on the overall performance.

In this evaluation, we use GloVe word representations to feed questions into the neural models, following the recommendations from the original papers for the models.
Such vector representations are based on co-occurrence statistics between the words and have been used extensively in the research community.
Recently, language modeling techniques such as BERT \cite{devlin2019BERTPretrainingDeep} have been shown to work well on the other types of clinical QA such as machine comprehension \cite{soni2020EvaluationDatasetSelection} as well as on the other clinical NLP tasks such as concept extraction \cite{fu2020ClinicalConceptExtraction}.
In the future, we aim to apply such techniques, that take into account the nuances of natural language via transfer learning, to the task of clinical SP.

\section{Conclusion}
\label{conclusion}

We applied different neural semantic parsers to the task of EHR QA to assess their performance relative to rule-based alternatives.
The performance of neural methods is promising given their ease of application and generalizability.
The performance disparity of the neural-based methods between different clinical datasets can be attributed to the diversity present in different corpora.
To understand the common sources of errors made by such neural models, we conducted a thorough error analysis using the incorrect predictions from all the neural models.

\vspace{-10pt}
\paragraph{Acknowledgments}
This work was supported by the U.S. National Library of Medicine, National Institutes of Health (NIH), (R00LM012104); the National Institute of Biomedical Imaging and Bioengineering (NIBIB: R21EB029575); the Cancer Prevention and Research Institute of Texas (CPRIT RP170668); and the UTHealth Innovation for Cancer Prevention Research Training Program Predoctoral Fellowship (CPRIT RP210042).

\bibliographystyle{vancouver-mod}
\setlength{\bibsep}{0.3pt}
\setlength{\itemsep}{0.8pt}
\bibliography{amia22}

\end{document}